\documentclass[10pt,twocolumn,letterpaper]{article}

\usepackage[pagenumbers]{cvpr} 

\definecolor{cvprblue}{rgb}{0.21,0.49,0.74}
\usepackage[pagebackref=cvprblue, breaklinks, colorlinks, citecolor=cvprblue]{hyperref}
\usepackage{multirow}
\usepackage{makecell}
\usepackage{multirow}
\usepackage{makecell}
\usepackage{bbding}
\usepackage{bbm}
\usepackage{amsmath}
\usepackage{amssymb}
\usepackage{colortbl}
\usepackage[table]{xcolor} 
\usepackage[linesnumbered,ruled,vlined]{algorithm2e}
\usepackage{xcolor}
\usepackage{textcomp,booktabs}
\usepackage{pifont}
\newcommand{\cmark}{\ding{51}}%
\newcommand{\xmark}{\ding{55}}%

\def\paperID{*****} 
\def\confName{CVPR}
\def\confYear{2025}

\title{When Person Re-Identification Meets Event Camera: A Benchmark Dataset and An Attribute-guided Re-Identification Framework}

\author{Xiao Wang$^{1}$, Qian Zhu$^{1}$, Shujuan Wu$^{1}$, Bo Jiang$^{1}$\thanks{Corresponding Author: Bo Jiang \& Shiliang Zhang}, Shiliang Zhang$^{2,3}$ \\  
${^1}${School of Computer Science and Technology, Anhui University, Hefei, China} \\ 
${^2}${Peng Cheng Laboratory, Shenzhen, China} \\ 
${^3}${School of Computer Science, Peking University, China} \\ 
\textit{\{zq542664, wushujuan0114\}@163.com}, \textit{\{xiaowang, jiangbo\}@ahu.edu.cn}, \\ \textit{\{slzhang.jdl\}@pku.edu.cn}  
}

\begin{document}
\maketitle

\begin{abstract}
Recent researchers have proposed using event cameras for person re-identification (ReID) due to their promising performance and better balance in terms of privacy protection, event camera-based person ReID has attracted significant attention. Currently, mainstream event-based person ReID algorithms primarily focus on fusing visible light and event stream, as well as preserving privacy. Although significant progress has been made, these methods are typically trained and evaluated on small-scale or simulated event camera datasets, making it difficult to assess their real identification performance and generalization ability. To address the issue of data scarcity, this paper introduces a large-scale RGB-event based person ReID dataset, called EvReID. The dataset contains 118,988 image pairs and covers 1200 pedestrian identities, with data collected across multiple seasons, scenes, and lighting conditions. We also evaluate 15 state-of-the-art person ReID algorithms, laying a solid foundation for future research in terms of both data and benchmarking. Based on our newly constructed dataset, this paper further proposes a pedestrian attribute-guided contrastive learning framework to enhance feature learning for person re-identification, termed TriPro-ReID. This framework not only effectively explores the visual features from both RGB frames and event streams, but also fully utilizes pedestrian attributes as mid-level semantic features.  Extensive experiments on the EvReID dataset and MARS datasets fully validated the effectiveness of our proposed RGB-Event person ReID framework. The benchmark dataset and source code will be released on \textcolor{red}{\url{https://github.com/Event-AHU/Neuromorphic_ReID}}
\end{abstract}

\section{Introduction} 

Person re-identification (ReID) is a critical research topic in the fields of computer vision and artificial intelligence. Its goal is to find pedestrians with the same ID as a given query sample from a set of candidate samples. It can be used in intelligent video surveillance, commercial and retail analysis, robotics and unmanned devices, etc. Most existing ReID algorithms are based on RGB cameras, making them highly susceptible to challenges such as illumination variations, motion blur, and privacy concerns.

\begin{figure}
\centering
\includegraphics[width=0.47\textwidth]{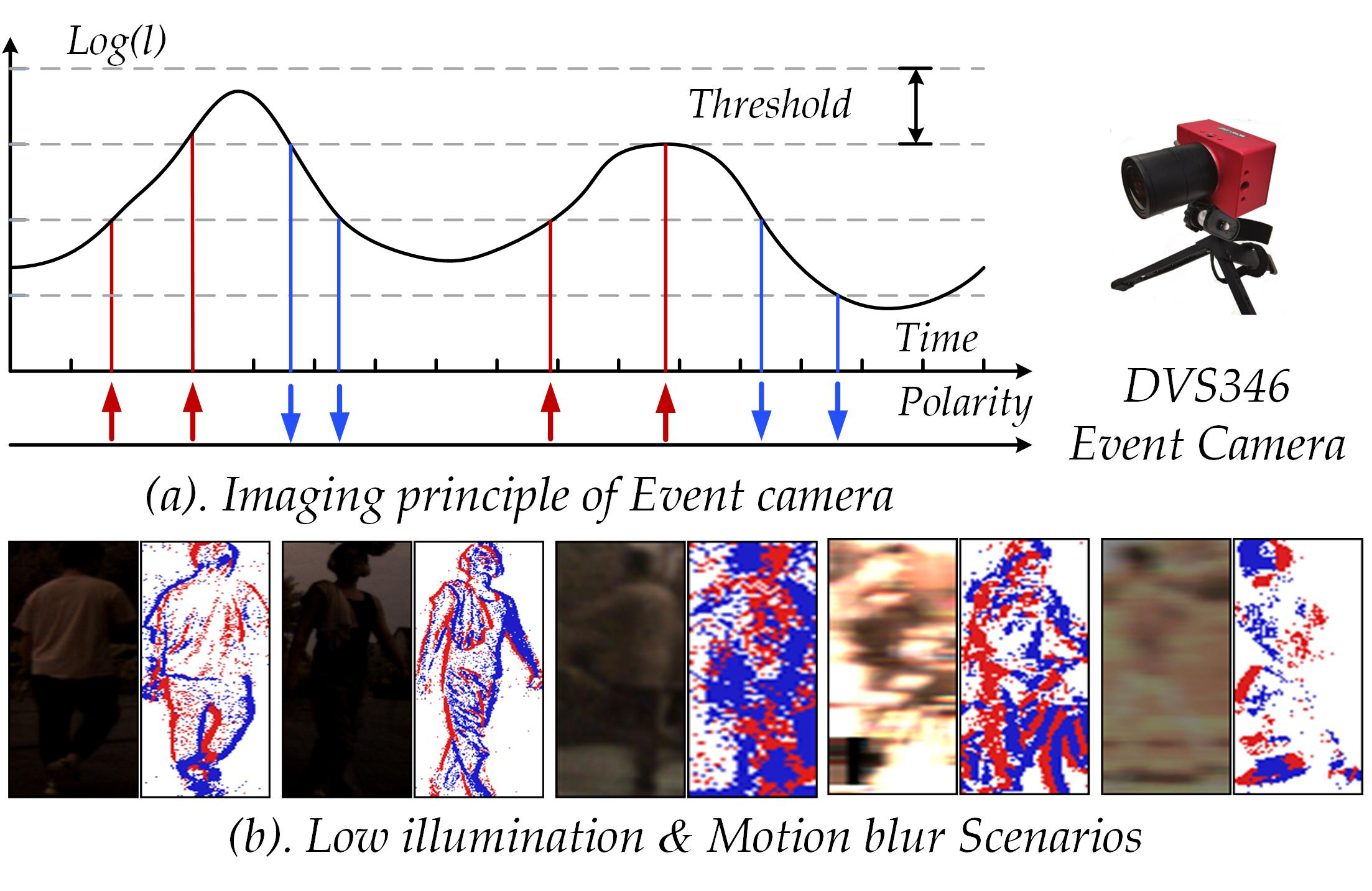} 
\caption{Imaging principle of event camera and comparison of RGB frames and event streams in challenging scenarios.} 
\label{IntroductionIMG}
\end{figure}

To address the aforementioned issues, some researchers resort to the bio-inspired event cameras for person re-identification~\cite{Ahmad_2023_ICCV, cao2023event, li2025event}, due to their advantages in low energy consumption, high dynamic range, no motion blur, and spatial sparsity~\cite{gallego2020eventSurvey}, as shown in Fig.~\ref{IntroductionIMG}. Specifically, Ahmad et al.~\cite{Ahmad_2023_ICCV} propose a person re-identification method through event anonymization, aiming to recognize individuals based on behavior and dynamic features without relying on identity information. Cao et al.~\cite{cao2023event} present an event-guided person re-identification method that leverages sparse-dense complementary learning, combining sparse event data and dense image features to enhance the robustness and accuracy of ReID across different scenarios. Li et al.~\cite{li2025event} propose a video person re-identification method that enhances performance by leveraging cross-modality fusion of visual and event data, along with temporal collaboration to capture dynamic motion in challenging scenarios.

Despite these progress, current event-based person re-identification algorithms are still limited by the following issues: 
1). Existing person ReID algorithms are typically trained and evaluated on small-scale or simulated event camera datasets, making it difficult to assess their real identification performance and generalization ability. 
2). Current ReID algorithms mainly focus on learning event stream features~\cite{li2025event, Ahmad_2023_ICCV} or fusing RGB and event features~\cite{cao2023event}, but fail to consider the semantic information, such as pedestrian attributes \textit{long hair, wearing glasses}~\cite{wang2022PARSurvey}, they can only achieve sub-optimal performance. Therefore, it is natural to raise the following question: ``\textit{How can we design a more effective and generalizable event-based person re-identification framework that leverages not only large-scale, real-world data but also incorporates rich semantic information such as pedestrian attributes?}"

Considering that the datasets for person re-identification based on event cameras are still scarce and limited in scale, for example, the Event ReID~\cite{Ahmad_2023_ICCV} dataset contains only 16,000 samples involving 33 pedestrians. In this paper, we first propose a new benchmark dataset for event stream-based person re-identification to bridge the data gap, termed EvReID. It contains 118,988 image pairs (7 times larger than the current real ReID dataset Event-ReID~\cite{Ahmad_2023_ICCV}) and covers 1200 pedestrian identities (36 times more than Event-ReID), with data collected across multiple seasons, scenes, and lighting conditions. We also evaluate 15 state-of-the-art person ReID algorithms, laying a solid foundation for future research in terms of both data and benchmarking. Some representative samples of our EvReID dataset are provided in Fig.~\ref{DatasetsPresentation}.

Based on our newly proposed benchmark dataset, we further propose a new pedestrian attribute-guided contrastive learning framework to enhance feature learning for person re-identification, termed TriPro-ReID. This framework not only effectively explores the visual features from both RGB frames and event streams, but also fully utilizes pedestrian attributes as mid-level semantic features. Specifically, given multi-modal input data, we first perform patching and projection to obtain RGB and event tokens, which are then fed into separate ViT backbone networks to learn spatio-temporal features. To facilitate feature interaction between the two modalities, we introduce a cross-modal prompt projector for cross-modal feature fusion. Meanwhile, we employ the  VTFPAR++~\cite{wang2024spatio} to predict pedestrian attributes from the input samples and use a text encoder to generate attribute semantic tokens. Subsequently, an attribute prompt injector is introduced to map these semantic tokens and fuse them with the visual features. We adopt widely-used metrics, including the triplet loss, ID loss, and the vision-attribute contrastive loss, to jointly guide the multi-modal feature learning for ReID. An overview of our proposed RGB-Event based person re-identification framework can be found in Fig.~\ref{TriPro}.

To sum up, the contributions of this paper can be summarized as the following three aspects: 

1). We propose a large-scale benchmark dataset for RGB-Event based person re-identification, termed EvReID dataset, which contains 118,988 image pairs and covers 1200 pedestrian identities, with data collected across multiple seasons, scenes, and lighting conditions. 15 state-of-the-art (SOTA) person ReID algorithms are re-trained and evaluated on this dataset, which lays a solid foundation for future research in terms of both data and benchmarking. 

2). We propose a pedestrian attribute guided contrastive learning framework for RGB-Event based person re-identification, termed TriPro-ReID. Both the semantic attribute and multi-modal visual features are exploited in our framework. 

3). Extensive experiments conducted on two benchmark datasets (i.e., EvReID and MARS dataset) fully validated the effectiveness of our proposed RGB-Event person ReID framework.

\textit{The following of this paper is organized as follows:} 
In Section~\ref{sec::relatedwork}, we review the related works on the event-based vision, person re-identification, and benchmark datasets. 
Then, we introduce our EvReID benchmark dataset in Section~\ref{sec::dataset} and describe the key modules of our framework in Section~\ref{sec::method}. After that, we conduct the experiments and give both quality and quantity analysis in Section~\ref{sec::experiments}. Finally, we conclude this paper and propose possible research directions in Section~\ref{sec::conclusion}.

\section{Related Works} \label{sec::relatedwork}

In this work, we propose to review the related works on Event-based Vision, Person Re-Identification, and Benchmark Datasets for Video-based Person Re-Identification. More works can be found in the following surveys~\cite{gallego2020eventSurvey}.

\subsection{Event-based Vision} 
Most computer vision systems rely on RGB modalities, which typically achieve superior performance. However, these systems tend to struggle in low-light environments. To address the shortcomings of RGB-based vision, numerous Event-based methods have been proposed, inspired by the unique pulse-like characteristics of events. In the re-identification field, Ahmad et al.~\cite{Ahmad_2023_ICCV} introduce the first Event-based person re-identification (ReID) dataset and develop an Event-anonymous network for privacy-preserving Event-based ReID. In the object detection field, RVT~\cite{gehrig2023recurrent}, GET~\cite{peng2023get}, and SAST~\cite{peng2024scene} achieve outstanding performance by leveraging a Transformer-based backbone, combined with an LSTM network and specialized attention mechanisms. MvHeat-DET~\cite{wang2024EvDET200K} uses a novel heat conduction-based foundational model to fully harness the advantages of events, achieving an optimal balance between accuracy and speed. SpikeYOLO~\cite{luo2024spikeYOLO}, SpikingYOLO~\cite{kim2020spikingyolo}, EMS-YOLO~\cite{su2023emsyolo} employ an SNN network to model Event stream data and integrate it with YOLO for object detection. In tracking field, Wang et al.~\cite{wang2024long}, Zhu et al.~\cite{zhu2024crsot} utilize RGB and Event dual-modal data for tracking and introduce a new multimodal tracking dataset. Mamba-FETrack~\cite{huang2024mamba} uses Mamba backbone networks to extract the features to realize more efficient tracking. However, even though Event-based methods address many of the limitations inherent in RGB cameras, RGB cameras still offer richer visual information, which Event cameras lack. This gap often results in Event-based methods achieving lower accuracy compared to their RGB counterparts. Therefore, we propose a dual-branch RGB-Event architecture to leverage the complementary strengths of both modalities, thereby improving performance.

\subsection{Person Re-Identification}  
Person re-identification (ReID)~\cite{shu2021large} has received increasing attention, particularly in video-based scenarios where temporal coherence between frames is exploited for more robust matching. 
Early methods typically adopted CNNs as backbones, such as AP3D~\cite{gu2020AP3D}, which utilizes 3D convolutions to jointly model spatial-temporal cues, and TCLNet~\cite{TCLNet}, which extracts complementary features through temporal modules. CTL~\cite{liu2021ctl}, PSTA~\cite{PSTA}, and STRF~\cite{aich2021strf} further enhance temporal modeling by integrating LSTM networks. Graph-based approaches have also been explored, with MGH~\cite{DBLP:conf/cvpr/YanQC0ZT020} introducing hypergraph structures to capture high-order spatial-temporal relations, and Chen et al.~\cite{chen2022keypoint} employing keypoint-based GCNs for refined motion representation. Meanwhile, Transformer-based architectures have shown strong potential for global context modeling. 
Works like DCCT~\cite{liu2023deeply}, VDT~\cite{zhang2024view}, PFD~\cite{wang2022pose}, and SBM~\cite{bai2022salient} adopt transformer backbones to improve representation quality. With the emergence of event cameras, methods such as~\cite{ahmad2024event, ahmad2023person, ahmad2022event, cao2023event} have explored event stream data to achieve high accuracy under challenging conditions while preserving privacy. Recently, the vision-language model CLIP~\cite{radford2021learning} has demonstrated strong transferability, prompting its adoption in ReID. Li et al.~\cite{li2023clip} introduce a two-stage training strategy to align image features with textual supervision, while SVLL-ReID~\cite{wang2025image} generates diverse textual queries to enhance CLIP’s matching capacity. TF-CLIP~\cite{tfclip} replaces static text with learned sequence representations, and CLIMB~\cite{yu2025climb} combines CLIP with the Mamba architecture to balance accuracy and efficiency. Several works further incorporate external semantic knowledge. IDEA~\cite{wang2025idea} generates structured captions for cross-modal consistency, and LATex~\cite{hu2025latex} integrates attribute-based prompts to bridge the aerial-ground domain gap. Despite these advances, most existing methods either ignore attribute semantics or rely on heavy models for semantic enhancement. In contrast, we propose an attribute-guided dual-modality ReID framework that leverages lightweight attribute recognition to predict semantic descriptions, encodes them into text features via a text encoder, and fuses them with visual representations. A cross-modal prompting mechanism is further introduced to enable effective interaction and mutual enhancement between modalities, resulting in fine-grained, semantically enriched representations for robust cross-modal ReID.

\subsection{Benchmark Datasets for Video-Based Person Re-Identification}  
The most widely used datasets for video-based person re-identification (ReID) include PRID-2011~\cite{hirzer2011person}, iLIDS-VID~\cite{wang2014person}, MARS~\cite{zheng2016mars}, DukeMTMC-VideoReID~\cite{wu2018exploit}, and LS-VID~\cite{li2019global}. PRID-2011 comprises 400 video sequences of 200 individuals captured by two static cameras, while iLIDS-VID contains 600 sequences of 300 pedestrians recorded under surveillance conditions from two disjoint camera views. As a large-scale benchmark, MARS is collected from six cameras, offering 17,503 tracklets for 1,261 identities along with 3,248 distractor sequences. DukeMTMC-VideoReID is a subset of the DukeMTMC~\cite{ristani2016performance} dataset, consisting of 702 identities for training, 702 for testing, and an additional 408 distractors. LS-VID further scales up video-based ReID with 3,772 annotated identities, offering more diverse visual scenarios.
In addition to RGB-based datasets, the event modality has been introduced into the ReID domain. Ahmad et al.~\cite{Ahmad_2023_ICCV} proposed the first event-based ReID dataset, Event-ReID, which contains event sequences for 33 individuals. However, existing datasets are either limited to a single modality or constrained by scale, and to date, no real-world person ReID dataset simultaneously includes both RGB and event modalities under a unified setting. To address this gap, we introduce EvReID, a new benchmark designed for dual-modality video-based ReID. It enables comprehensive evaluation of RGB, event, and RGB-Event fusion methods under consistent spatio-temporal conditions, and serves as a foundation for developing and benchmarking future multi-modal person ReID approaches.

\begin{table*}[!htp]
\centering
\caption{Statistics of existing video-based person ReID datasets. \textbf{Seasons} indicates the diversity of data collection seasons, \textbf{Lighting} refers to the time periods of video capture, \textbf{Events-Reality} denotes whether the event data were acquired from real event sensors.} 
\label{bench_dataset}
\resizebox{\textwidth}{!}{ 
\begin{tabular}{c|l|c|c|c|c|c|c|c}
\hline \toprule[0.5pt]
\textbf{\#Index} &\textbf{Dataset} &\textbf{Year} & \textbf{\#IDs} & \textbf{\#Images} & \textbf{Modality} & \textbf{Seasons} & \textbf{Lighting}  & \textbf{Real Event}\\
\hline
01 & \textbf{PRID-2011}~\cite{hirzer2011person}            & 2011 & 200    & 40,033     & RGB        & Single & Daytime  &\xmark \\
02 & \textbf{SAIVT-Softbio}~\cite{bialkowski2012database}  & 2012 & 152  & 64,472     & RGB        & Single & Daytime &\xmark \\
03 & \textbf{iLIDS-VID}~\cite{wang2014person}              & 2014 & 300    & 42,460     & RGB        & Single & Daytime  &\xmark \\
04 & \textbf{HDA Person}~\cite{nambiar2014multi}           & 2014 & 53  & 2,976      & RGB        & Single & Daytime &\xmark \\
05 & \textbf{MARS}~\cite{zheng2016mars}                    & 2016 & 1,261  & 1,067,516  & RGB        & Single & Daytime &\xmark \\
06 & \textbf{DukeMTMC-VideoReID}~\cite{wu2018exploit}    & 2018 & 1,404 & 815,420    & RGB        & Single & Daytime  &\xmark \\
07 & \textbf{LS-VID}~\cite{li2019global}                   & 2019 & 3,772 & 2,982,685  & RGB        & Single & \textbf{Day, Night}  &\xmark  \\ 
08 & \textbf{Event ReID}~\cite{Ahmad_2023_ICCV}            & 2023 & 33     & 16,000     & Event      & Single & Daytime & \cmark\\
\hline
09 & \textbf{EvReID (Ours)}                                &2025 & 1,200  & 118,988    & \textbf{RGB-Event} & \textbf{Multi} & \textbf{Day, Night} &\cmark \\
\hline \toprule[0.5pt]
\end{tabular}} 
\end{table*}

\section{EvReID Benchmark dataset} \label{sec::dataset}

\subsection{Protocols} 
To provide a good platform for the training and evaluation of RGB-Event person re-identification, we construct the EvReID benchmark dataset. When collecting data for the EvReID dataset, we obey the following protocols: 
\textit{1). Large Scale:} EvReID is the first video-based Re-ID dataset, which integrates both RGB and Event modalities. It includes 1,200 unique identities, captured over 118,988 frames, providing a comprehensive foundation for training and evaluation.
\textit{2). Multiple Viewpoints:} We use a DVS346 Event camera to capture spatiotemporally aligned RGB-Event sample pairs. Specifically, by following the direction of the person's movement and rotating the camera's viewpoint, we obtain images of the same pedestrian from different viewpoints.  
\textit{3). Complex and Varied Scenes:} The dataset reflects real-world variability by incorporating pedestrian footage across different times of day, seasons, and weather conditions, ensuring it covers a wide range of scenarios that pedestrians encounter in practice.
\textit{4). Bi-modal Complementarity:} We add 11 kinds of different noises on the RGB modal of the EvReID dataset, which include light change, motion blur, and adverse weather conditions, to validate complementary learning for enhanced bi-modality.

\begin{figure*}
\centering
\includegraphics[width=0.98\textwidth]{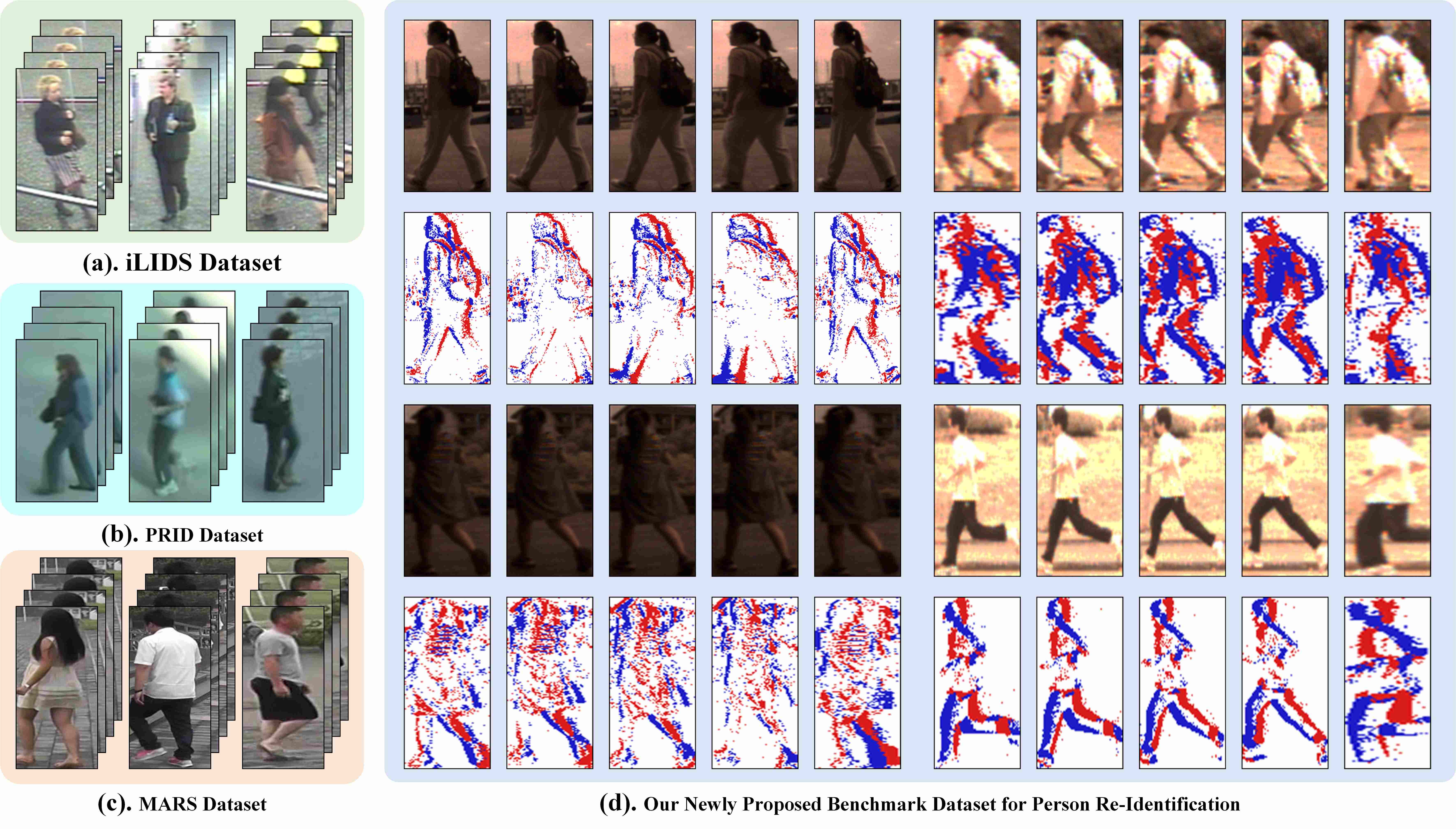} 
\caption{Comparison of existing video-based ReID datasets and our newly proposed EvReID dataset.}
\label{DatasetsPresentation}
\end{figure*}

\subsection{Data Collection and Statistical Analysis} 

As shown in Table~\ref{bench_dataset}, we compare several representative video-based person re-identification (ReID) datasets with our proposed EvReID dataset. Most existing benchmarks are based solely on the RGB modality, which performs well under well-lit conditions but struggles to capture rapid motion or subtle visual cues in low-light or nighttime environments. To address these limitations, recent efforts have attempted to simulate event-based reID by generating pseudo-event data from RGB frames. 
Notably, Ahmad et al.~\cite{Ahmad_2023_ICCV} proposed the first real-world Event-based ReID dataset, emphasizing privacy protection and robustness in complex scenes. However, their dataset contains only 33 identities, which significantly limits its scalability and suitability for large-scale ReID research. Moreover, many existing datasets are collected in overly constrained or homogeneous environments, limiting their generalization to real-world deployment scenarios.

In contrast, our proposed EvReID dataset is the first large-scale benchmark specifically designed for both Event-based and RGB-Event (RGBE) person ReID tasks. As illustrated in Fig.~\ref{DatasetsPresentation}, compared with mainstream video-based person ReID datasets we show in the first row, EvReID provides diverse samples across multiple seasons, lighting conditions (including daytime and nighttime), and weather variations, making it significantly more representative of real-world applications. The dataset is collected using the DVS346 Event camera at a resolution of $346 \times 260$ and comprises {118,988 frame pairs from 1,200 distinct identities}. Event streams are processed into Event frames and are spatiotemporally aligned with corresponding RGB frames, ensuring consistent temporal correspondence across modalities. To ensure fair evaluation, the dataset is split into non-overlapping training and test sets with a 70\%/30\% split. For evaluation, we adopt a {single-shot setting}, where the query and gallery sets are strictly separated.

\subsection{Benchmark Baselines} 
To establish a comprehensive benchmark dataset for RGB-Event based person re-identification, we select 15 SOTA or representative methods for evaluation on our proposed dataset, including:
\textbf{1). CNN-based:} OSNet~\cite{zhou2019omniscalefeaturelearningperson}, TCLNet~\cite{TCLNet}, AP3D~\cite{gu2020AP3D}, MGH~\cite{DBLP:conf/cvpr/YanQC0ZT020}, STMN~\cite{eom2021video}, PSTA~\cite{PSTA}, GRL~\cite{liu2021watching}, BiCnet-TKS~\cite{BiCnet-TKS}, SINet~\cite{bai2022SINet}, and SDCL~\cite{cao2023event}.
\textbf{2). Transformer-based:} DCCT~\cite{liu2023deeply}, CLIP-ReID~\cite{li2022clip}, TF-CLIP~\cite{tfclip}, DeMo~\cite{wang2025decoupled} and CLIMB-ReID~\cite{yu2025climb}.

To ensure a fair and unified evaluation across modalities, each baseline model is extended to support both RGB and Event stream inputs. This is achieved by duplicating the original single-stream architecture into two parallel branches, each responsible for processing one specific modality. Both branches share the same network structure and parameter configuration as the original design. After extracting features from the two modalities, the outputs are concatenated at the embedding level to generate a joint representation for person re-identification. This fusion strategy preserves the integrity of the original models and introduces no structural modifications, ensuring that performance comparisons remain focused on the influence of multimodal input rather than architectural differences.

\section{Methodology} \label{sec::method}

\subsection{Preliminary: CLIP and CLIP-ReID}
CLIP~\cite{radford2021learning} consists of a visual encoder $\mathcal{V(\cdot)}$ and a text encoder $\mathcal{T(\cdot)}$, which are jointly trained using contrastive learning to project input images and texts into a shared representation space. Specifically, let ${\{img_i, text_i\}_{i = 1}^B}$ denote a batch of $B$ paired visual-language training samples, where $\{img_i\}$ is an image and $\{text_i\}$ is the corresponding textual description. CLIP employs the two encoders along with two linear projection layers to encode images and texts into corresponding feature embeddings. The similarity between image and text features is computed as:
\begin{equation}
    \label{preliminaries}
    S(img_i, text_i) = \mathcal{J}(\mathcal{V}(img_i)) \cdot \mathcal{J}(\mathcal{T}(text_i))
\end{equation}
where $\mathcal{J}_v$ and $\mathcal{J}_t$ are linear layers that project features into a unified embedding space. Two contrastive losses, denoted as $L{v2t}$ and $L{t2v}$, are employed to align visual and textual features during training. In downstream tasks, the textual input is typically formulated as prompts such as “A photo/video of a \{class\},” where “\{\}” is replaced with a specific class label (\textit{e.g.}, cat). However, in person ReID, identity annotations are merely represented as integer indices rather than meaningful text.
To bridge this gap, CLIP-ReID~\cite{li2023clip} introduces learnable identity-specific tokens to construct personalized textual prompts. These prompts take the form of “A photo of a $[X]_1, [X]_2, \cdots, [X]_{Num}$ person”. During an additional training stage, the parameters of $\mathcal{V}(\cdot)$ and $\mathcal{T}(\cdot)$ are frozen, and the identity tokens are optimized using $L{v2t}$ and $L_{t2v}$ to learn a discriminative but implicit textual representation for each identity. While CLIP-ReID achieves promising results on conventional RGB-based person ReID datasets, a simple dual-branch CLIP-ReID architecture proves insufficient when applied to RGB-Event dual-modality scenarios. It fails to fully leverage the complementary temporal and modality-specific cues provided by the Event stream. More critically, text prompts such as “A photo of a $[X]_1, [X]_2, \cdots, [X]_{Num}$ person”, though identity-aware, do not explicitly model or utilize the fine-grained pedestrian attribute information inherently present in video frames.

\subsection{Overview} 
We propose a novel framework, \textbf{Tri}ple-\textbf{Pro}mpting \textbf{Re-id}entification (TriPro-ReID), which consists of two core modules: Positive-Negative Attribute Prompting (PNAP) and Cross-Modal Prompting (CMP). The framework is optimized through a three-stage training strategy. In the first stage, we leverage the vanilla CLIP contrastive learning pipeline, initializing identity-aware context prompts for each person's identity and associating them with corresponding IDs to obtain their textual representations. In the second stage, we fix the identity-aware context prompts and introduce visual prompts into the visual encoder, which helps align visual features with textual representations. Additionally, we introduce CMPs to bind the visual features from both RGB and Event modalities with the identity embeddings, and use the cross-modality prompt propagation to convey multimodal complementary information. This cross-modal alignment technique can effectively utilise complementary information from different modalities and extract generalized information from pre-trained CLIP models. In the third stage, we incorporate PNAPs, which are generated based on person attributes and encoded with the text encoder. These PNAPs are injected into the intermediate layers of the visual encoder to dynamically modulate the visual features with attribute-aware discrimination. The model is then supervised using ID and triplet losses to optimize it for person ReID tasks. The following sections will provide a detailed introduction to these modules.

\begin{figure*}[!htp]
\centering
\includegraphics[width=\textwidth]{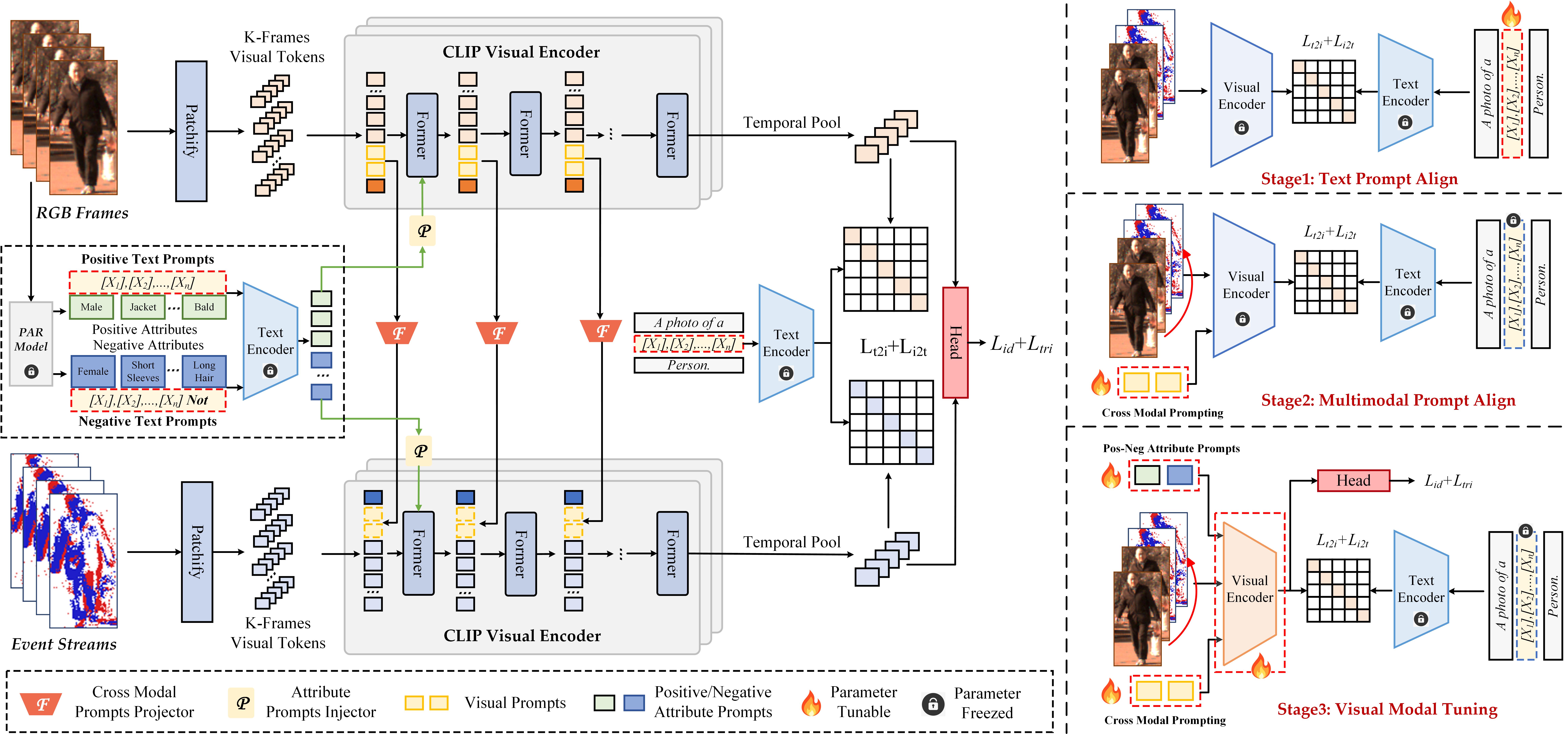}
\caption{An illustration of our proposed TriProReId framework. The left part illustrates the overall architecture of the RGB-Event person re-identification framework, consisting of dual-stream backbones and multi-level fusion modules. The right part presents the three-stage training strategy, including {Text Prompt Align}, {Multimodal Prompt Align}, and {Visual Modal Tuning} stages.}
\label{TriPro}
\end{figure*}

\subsection{Input Representation}  
Given the video-based person ReID training sets $\mathcal{D}_R = \{(x_i, y_i)\}_{i=1}^{N_s}$ and $\mathcal{D}_E = \{(x_i, y_i)\}_{i=1}^{N_s}$ in the RGB and Event modalities, respectively, where $y_i \in \{1, \cdots, Y\}$ denotes the identity label and $N_s$ is the total number of person tracklets in each modality, we aim to extract identity-specific sequence representations. 
Taking an identity $y_i$ as an example, we first employ the pre-trained CLIP visual encoder to extract sequence-level features for all RGB and Event video sequences corresponding to $y_i$, denoted as $\hat{F}_{R}$ and $\hat{F}_{E}$, respectively.
Specifically, considering the RGB modality, we take an image sequence $\{I_{R_t}\}_{t=1}^T$ of $T$ frames belonging to identity $y_i$ as input. Each frame $I_{R_t} \in \mathbb{R}^{H \times W \times 3}$ is divided into $N_p$ non-overlapping patches $I_{R_t,i} \in \mathbb{R}^{P^2 \times 3}$, where $H$ and $W$ denote the height and width of the image, and $P$ is the patch size. The number of patches is given by $N_p = \frac{H}{P} \times \frac{W}{P}$. 
Each patch is projected into token embeddings via a linear projection layer $E_{\text{emb}} \in \mathbb{R}^{3P^2 \times D}$, where $D$ is the token dimension. Additionally, a learnable class token $\mathrm{[CLS]}$ for each frame, denoted as $CLS_{R_t} \in \mathbb{R}^D$, is prepended to the sequence of patch embeddings. The resulting input to the visual encoder at frame $t$ is formulated as:
\begin{equation}
    Z_{R_t}^{\text{Patch}} = [CLS_t, E_{\text{emb}} I_{t, 1}, \cdots, E_{\text{emb}} I_{t, N_p}] + e^{\text{sp}}
\end{equation}
where $e^{\text{sp}}$ denotes the spatial positional embedding. These embeddings are subsequently processed by the CLIP visual encoder. The resulting frame-level feature representation is:
\begin{equation}
    \hat{F}_{R} = \mathcal{V}(Z_{R_t}^{\text{Patch}})
\end{equation}
where $\mathcal{V}(\cdot)$ denotes the CLIP visual encoder.
To incorporate identity-level textual guidance, we construct an ID-specific prompt: “A photo of a $[X]_1, [X]_2, \cdots, [X]_{\text{Num}}$ person”, where each $[X]_{y_i}$ ($y_i \in \{1, \dots, \text{Num}\}$) is a learnable token with the same dimension as the word embedding and is independently learned for each identity. The textual prompt is fed into the CLIP text encoder $\mathcal{T}(\cdot)$ to obtain the identity-specific text feature:
\begin{equation}
    T_{y_i} = \mathcal{T}(\text{text}_{y_i})
\end{equation}
where $\text{text}_{y_i}$ is the generated textual description for identity $y_i$.

\subsection{Attribute Guided Prompt Learning} 
We design a dual-stream network architecture for RGB-Event person ReID, which leverages CLIP-ReID as our baseline and introduces two main modules to enhance feature fusion between RGB and Event modalities and explore attribute semantic information for robust person ReID.

\noindent $\bullet$ \textbf{Positive-Negative Attribute Prompts.} 
To further enhance the discriminative capacity of identity representations with fine-grained semantic cues, we propose a \textit{Positive-Negative Attribute Prompting} (PNAP) mechanism. This design is inspired by the learnable context prompt strategy in~\cite{zhou2022learning, khattak2023maple}, and is further motivated by~\cite{wang2025idea}, which demonstrates the benefits of incorporating attribute-level semantic priors for multi-modal object re-identification. Specifically, we employ a pre-trained pedestrian attribute recognition model~\cite{wang2024spatio}, denoted as $\mathcal{PAR}(\cdot)$, to predict attribute labels for a given input sequence. Based on the predicted results, we construct a \textit{positive attribute prompt} $A^{\text{pos}}$, which contains all the predicted attributes (e.g., “\textit{Male, Jacket, Bald}”), and a \textit{negative attribute prompt} $A^{\text{neg}}$ by negating the unpredicted attributes (e.g., “\textit{Not Female, Not Short Sleeves, Not Long Hair}”).
Both types of prompts are encoded into textual embeddings using the frozen CLIP text encoder $\mathcal{T}(\cdot)$. Following the prompt tuning paradigm of CoOp~\cite{zhou2022learning}, we further introduce a set of learnable context vectors: $P^{\text{pos}}$ for positive prompts and $P^{\text{neg}}$ for negative ones. These are used to adapt the attribute prompts to the target person ReID task. The enriched prompts are then passed through a fully connected (FC) layer and concatenated with the visual features $\hat{F}_R$ and $\hat{F}_E$ to yield prompt-enhanced identity representations $F_R^0$ and $F_E^0$. The overall process can be summarized as:
\begin{align}
    &\mathcal{A} = \mathcal{PAR}(I_R) \\
    & PNAP = \mathcal{T}(\mathcal{A}^{\text{pos}}, P^{\text{pos}}) \cup  \mathcal{T}(\mathcal{A}^{\text{neg}}, P^{\text{neg}}) \\
    & F_{R}^0 = \mathrm{Con}(\hat{F}_R, \mathrm{FC}(PNAP)) \\
    & F_{E}^0 = \mathrm{Con}(\hat{F}_E, \mathrm{FC}(PNAP))
\end{align}
Here, $\mathcal{PAR}(\cdot)$ is the attribute recognition model, $\mathcal{T}(\cdot)$ denotes the CLIP text encoder, and $\mathrm{Con}(\cdot)$ represents the feature concatenation operation. $F_R^0$ and $F_E^0$ serve as the prompt-augmented input representations for the RGB and Event branches, respectively.

\noindent $\bullet$ \textbf{Cross Modal Prompts.} Inspired by the Visual Prompt Tuning paradigm~\cite{jia2022visual}, we design a Cross Modal Prompting (CMP) mechanism to facilitate bidirectional interaction between RGB and Event modalities at the feature level, which introduces a set of learnable prompts to enable early-stage alignment across modalities. Concretely, we initialize a set of modality-specific prompt tokens $CMP^0_R$ for the RGB branch and project them into the Event branch via an FC layer to obtain $CMP^0_E$. These prompts are concatenated with the visual features $F_R^0$ and $F_E^0$, resulting in the prompt-injected features $F_R^1$ and $F_E^1$, respectively:
\begin{align}
    &F_{R}^1 = \mathrm{Con}(F_{R}^0, CMP^0_{R}) \\
    &CMP^0_{E} = \mathrm{FC}(CMP^0_{R}) \\
    &F_{E}^1 = \mathrm{Con}(F_{E}^0, CMP^0_{E})
\end{align}
Here, $F_R^0$ and $F_E^0$ denote the visual features at the input (i.e., layer 0) of the transformer for the RGB and Event branches, respectively, while $CMP_R^0$ and $CMP_E^0$ represent the initial Cross Modal Prompt tokens for each modality. As the features propagate through the transformer layers, the RGB-side CMPs are iteratively updated to $CMP_R^1, \dots, CMP_R^k$, where the superscript $k$ indicates the corresponding transformer layer. At each layer, the CMPs in the Event branch are synchronously replaced with the transformed RGB-side CMPs from the same layer, enabling continuous and guided feature fusion between the two modalities throughout the network.

\subsection{Training Strategy} 
To effectively bridge the modality gap and enhance discriminative representation learning for RGB-Event person re-identification, we propose a three-stage training strategy, progressively aligning text, vision, and multimodal cues. The training stages are illustrated in Fig.~\ref{TriPro} and described as follows.

\noindent $\bullet$ \textbf{Stage 1: Text Prompt Alignment.}
In the first stage, we follow the paradigm of CLIP-ReID and introduce ID-specific learnable prompts to model ambiguous identity-related semantics. For each identity, we learn a set of $n$ text tokens ${[X]_1, [X]_2, \cdots, [X]_n}$, which are concatenated to form a textual description in the format of “A photo of a $[X]_1, [X]_2, \cdots, [X]_n$ person”. These prompts are optimized while keeping both the visual encoder $\mathcal{V}(\cdot)$ and text encoder $\mathcal{T}(\cdot)$ frozen. The optimization is guided by two symmetric contrastive losses, vision-to-text ($L{v2t}$) and text-to-vision ($L{t2v}$), which encourage cross-modal alignment between video features and the learnable textual embeddings. Formally, given a mini-batch of size $|B|$, the overall objective is defined as:
\begin{align}
    &L_{v2t} = -\frac{1}{|B|} \sum_{i=1}^{|B|} \log \frac{\exp({v}_i^\top {t}_i / \tau)}{\sum_{j=1}^{|B|} \exp(v_i^\top {t}_j / \tau)} \\
    &L_{t2v} = -\frac{1}{|B|} \sum_{i=1}^{|B|} \log \frac{\exp({t}_i^\top {v}_i / \tau)}{\sum_{j=1}^{|B|} \exp(t_i^\top {v}_j / \tau)} \\
    &L_{stage1} = L_{v2t} + L_{t2v}
\end{align}
where ${v}_i$ and ${t}_i$ represent the visual and text features of the $i$-th sample, $\tau$ is a temperature hyperparameter, $L_{v2t}$ and $L_{t2v}$ encourage the visual features to match their paired textual descriptions in the embedding space. This stage helps to warm up the model with fundamental visual-language alignment before introducing multimodal and fine-grained supervision.

\noindent $\bullet$ \textbf{Stage 2: Multimodal Prompt Alignment.}
To integrate complementary information from both RGB and Event streams, we introduce learnable CMP in the second stage. These prompts serve as a bridge between modalities, enabling the model to capture and fuse heterogeneous visual patterns from both input types.
Specifically, we concatenate the CMP tokens with the patch embeddings of RGB and Event frames, forming a unified multimodal token sequence. This sequence is then input to the frozen visual encoder $\mathcal{V}(\cdot)$, allowing the CMPs to guide the fusion process during token interactions. The text encoder $\mathcal{T}(\cdot)$ remains frozen, and we continue to apply the contrastive loss:
\begin{equation}
    L_{stage2} = L_{v2t} + L_{t2v}
\end{equation}
This stage focuses purely on visual modality fusion, enhancing the joint representation of RGB and Event data through CMP while retaining language guidance established in Stage 1.

\begin{algorithm}[t]
\caption{Training Procedure of TriPro-ReID}
\label{TriProAlgorithm}
\KwIn{Train set $\mathcal{D}$, attributes for each identity, and CLIP encoders $\mathcal{V}$ and $\mathcal{T}$}
\vspace{0.5em}
\textbf{Stage 1: Text Prompt Alignment}\;
Freeze $\mathcal{V}$ and $\mathcal{T}$; Update ID-specific learnable text prompts\;
\For{each batch $(I_{R,t}, I_{E,t}, y)$ in $\mathcal{D}$}{
    Concatenate the multimodal visual features in channel dimension, and apply temporal pooling to obtain final representations\;
    Compute contrastive loss $L_{v2t} + L_{t2v}$\;
}
\vspace{0.5em}
\textbf{Stage 2: Multimodal Prompt Alignment}\;
Keep $\mathcal{V}$ and $\mathcal{T}$ frozen; initialize cross-modal prompts (CMPs)\;
\For{each batch $(I_{R,t}, I_{E,t}, y)$ in $\mathcal{D}$}{
    Inject CMP tokens into patch embeddings\;
    Concatenate and input into $\mathcal{V}$ to obtain joint features\;
    Compute $L_{v2t} + L_{t2v}$\;
}
\vspace{0.5em}
\textbf{Stage 3: Visual-Modal Fine-Tuning with Attribute Prompts}\;
Unfreeze $\mathcal{V}$; Encode PNAPs via $\mathcal{T}$\;
\For{each batch $(I_{R,t}, I_{E,t}, y)$ in $\mathcal{D}$}{
    Inject PNAPs into $\mathcal{V}$ via attribute prompter\;
    Concatenate CMP, PNAP, and patch embeddings\;
    Compute full loss: $L_{id} + L_{tri} + L_{v2t} + L_{t2v}$\;
}
\end{algorithm}

\noindent $\bullet$ \textbf{Stage 3: Visual-Modal Tuning with Attribute Prompts.}
In the final stage, we fine-tune the visual encoder by injecting PNAP to enhance attribute-level discrimination. These prompts are generated based on pre-estimated person attributes (e.g., “Male”, “Bald”, “Not Long Hair”) and encoded using the text encoder $\mathcal{T}(\cdot)$.
The PNAP tokens are injected into intermediate layers of the visual transformer via a dedicated Attribute Prompt Injector, allowing the model to modulate visual features dynamically according to semantic cues. Additionally, both CMP and standard learnable visual prompts are preserved and fused at this stage, enabling simultaneous attribute awareness and modality integration.
After temporal pooling over the fused visual tokens, the final feature is supervised by a classification head using ID and triplet losses:
\begin{align}
    &L_{id} = \frac{1}{|B|} \sum_{i=1}^{|B|} \mathrm{CrossEntropy}({v}_i, {y}_i) \\
    &L_{tri} = \frac{1}{|B|} \sum_{i=1}^{|B|} \max(0, \| {v}_i - {v}_+ \|_2^2 - \|{v}_i - {v}_- \|_2^2 + \alpha) \\
    &L_{stage3} = L_{id} + L_{tri} + L_{v2t} + L_{t2v}
\end{align}
where $|B|$ represents the batch size, ${v}_i$ denotes the feature vector of the $i$-th sample, ${y}_i$ is the true identity label, ${v}_+$ and ${v}_-$ are the feature vectors of the positive and negative samples, respectively, $\|\cdot\|_2$ denotes the L2 norm (Euclidean distance), and $\alpha$ is the margin used in the triplet loss to ensure sufficient distance between positive and negative samples. The terms $L_{id}$, $L_{tri}$, $L_{v2t}$, and $L_{t2v}$ together define the overall loss function, where $L_{id}$ enforces identity classification, $L_{tri}$ encourages correct relative distances between features, and $L_{v2t}$ and $L_{t2v}$ ensure cross-modal alignment between visual and textual features.

The whole training process of the proposed TriPro-ReID is summarized in Algorithm~\ref{TriProAlgorithm}. We first leverage learnable contextual prompts to mine and store the hidden states of the pre-trained visual encoder and text encoder, allowing CLIP to retain its own advantages. During the second stage, the cross modal prompts are utilized to learn effective RGB-Event fusion. In the final stage, the positive-negative attribute prompts are joined to enhance attribute-aware discrimination.

\section{Experiments} \label{sec::experiments}

\begin{table*}[htp]
\center
\small  
\caption{Comparison with public methods on our datasets. The highest results are shown in \textbf{bold}, while the second-best results are indicated with \underline{underline}.} 
\label{ComparisononEvReIDdataset}
\begin{tabular}{l|c|l|cccc}
\hline \toprule [0.5 pt] 
\multirow{1}{*}{\textbf{Methods}} & \multirow{1}{*}{\textbf{Modality}} & \multirow{1}{*}{\textbf{Publish}} & \multirow{1}{*}{\textbf{mAP}} & \multirow{1}{*}{\textbf{Rank-1}} & \multirow{1}{*}{\textbf{Rank-5}} & \multirow{1}{*}{\textbf{Rank-10}} \\  
\hline
\#01 AP3D~\cite{gu2020AP3D}  & V & ECCV$_{2020}$   & 65.4  & 83.0 & 92.1 & 95.3 \\
\#02 PSTA~\cite{PSTA}  & V & ICCV$_{2021}$   &  63.3  & 85.2 & 92.5 & 95.3 \\
\#03 GRL~\cite{liu2021watching}  & V & CVPR$_{2021}$ & 34.5  & 58.2 & 73.6 & 84.0 \\
\#04 SINet~\cite{bai2022SINet}  & V  & CVPR$_{2022}$  & 62.7  & 83.0 & 92.4 & \textbf{96.5} \\
\#05 CLIP-ReID~\cite{li2022clip} & V & AAAI$_{2023}$  & 49.9 & 68.8 & 81.0 & 84.0 \\
\hline
\#01 AP3D~\cite{gu2020AP3D}  & E & ECCV$_{2020}$ & 40.6  & 67.3 & 80.8 & 86.2 \\
\#02 PSTA~\cite{PSTA}  & E & ICCV$_{2021}$  &  37.1  & 61.0 & 77.4 & 84.6 \\
\#03 GRL~\cite{liu2021watching}  & E & CVPR$_{2021}$  & 38.9 & 62.6 & 78.0 & 83.6 \\
\#04 SINet~\cite{bai2022SINet}  & E  & CVPR$_{2022}$  & 40.1  & 67.0 & 81.5 & 85.2 \\
\#05 CLIP-ReID~\cite{li2022clip} & E & AAAI$_{2023}$ & 30.4 & 52.5 & 70.3 & 79.1 \\
\hline

\#01 OSNet~\cite{zhou2019omniscalefeaturelearningperson}  & V+E  & ICCV$_{2019}$  & 23.7  & 49.1 & 65.4 & 72.3 \\
\#02 TCLNet~\cite{TCLNet}  & V+E  & ECCV$_{2020}$   & 55.8  & 77.4 & 89.0 & 93.1 \\
\#03 AP3D~\cite{gu2020AP3D}  & V+E & ECCV$_{2020}$ & 66.9  & \underline{86.5} & \textbf{95.6} & \textbf{96.5} \\
\#04 MGH~\cite{DBLP:conf/cvpr/YanQC0ZT020}  & V+E & CVPR$_{2020}$   & 43.2  & 70.9 & 89.4 & 92.7 \\
\#05 STMN~\cite{eom2021video}  & V+E & ICCV$_{2021}$  & 42.1  & 73.8 & - & - \\ 
\#06 PSTA~\cite{PSTA}  & V+E & ICCV$_{2021}$   &  68.2  & 82.3 & 90.8 & 94.5 \\
\#07 GRL~\cite{liu2021watching}  & V+E & CVPR$_{2021}$  & 38.9  & 62.6 & 78.0 & 83.6 \\
\#08 BiCnet-TKS~\cite{BiCnet-TKS}  & V+E & CVPR$_{2021}$  & 50.8  & 80.5 & 89.6 & 92.5 \\ 
\#09 SINet~\cite{bai2022SINet}  & V+E  & CVPR$_{2022}$  & 67.1  & 83.4 & 93.0 & 95.6 \\
\#10 DCCT~\cite{liu2023deeply}  & V+E & TNNLS$_{2023}$  & 24.6  & 42.7 & 64.9 & 75.5 \\
\#11 CLIP-ReID~\cite{li2022clip}  & V+E & AAAI$_{2023}$   & 49.2 & 73.0 & 85.5 & 91.5 \\
\#12 SDCL~\cite{cao2023event}   & V+E & CVPR$_{2023}$  & 54.2  & 69.3 & 83.8 & 87.1 \\
\#13 TF-CLIP~\cite{tfclip}  & V+E & AAAI$_{2024}$  & 56.9  & 78.6 & 91.8 & 94.3 \\
\#14 DeMo~\cite{wang2025decoupled}  & V+E & AAAI$_{2025}$   & 59.4 & 75.7 & 90.1 & 92.8 \\
\#15 CLIMB-ReID~\cite{yu2025climb}  & V+E & AAAI$_{2025}$   & \underline{68.3} & 85.2 & 92.8 & \underline{95.8} \\
\hline
\#16 TriPro-ReID (Ours)  & V+E & -  &  \textbf{69.3} &  \textbf{88.6}   & \underline{94.3}   & 95.4 \\ 
\hline \toprule [0.5 pt] 
\end{tabular}
\end{table*}

\begin{table}[!htb]
\caption{Comparison with SOTA methods on MARS$^*$ dataset. The highest results are shown in \textbf{bold}, while the second-best results are indicated with \underline{underline}.} 
\label{Comparisononpublicdatasets} 
\centering
\small
\resizebox{0.45\textwidth}{!}{
\begin{tabular}{l|c|cc}
\hline \toprule [0.5 pt] 
\multirow{1}{*}{\textbf{Methods}} & \multirow{1}{*}{\textbf{Modality}} & \multirow{1}{*}{\textbf{mAP}} & \multirow{1}{*}{\textbf{Rank-1}} \\

\hline
\#01 OSNet~\cite{zhou2019omniscalefeaturelearningperson}  & V  & 81.4  & 87.3 \\
\#02 GRL~\cite{liu2021watching}  & V & 84.8  & 91.0  \\
\#03 STMN~\cite{eom2021video}  & V & 84.5  & 90.5 \\
\#04 PSTA~\cite{PSTA} & V  & 85.8  & 91.5 \\
\#05 TF-CLIP~\cite{tfclip} & V & \underline{89.4}  & \underline{93.0} \\
\#06 CLIMB-ReID~\cite{yu2025climb}  & V & \textbf{89.7}   & \textbf{93.3} \\

\hline
\#01 TCLNet~\cite{TCLNet}  & E & 38.2 & 25.3 \\
\#02 GRL~\cite{liu2021watching}  & E & 27.7  & 16.7  \\
\#03 BiCnet-TKS~\cite{BiCnet-TKS}  & E & 30.9 & 17.3 \\
\#04 STMN~\cite{eom2021video}  & E & 22.4  & 10.0 \\

\hline
\#01 OSNet~\cite{zhou2019omniscalefeaturelearningperson} & V+E & 81.9 & 87.7  \\			
\#02 GRL~\cite{liu2021watching} & V+E & 82.8 & 88.7   \\
\#03 SRS-Net~\cite{wang2020simple} & V+E & 83.8 & 89.3  \\ 				
\#04 STMN~\cite{eom2021video} & V+E & 83.4 & 89.0 \\
\#05 PSTA~\cite{PSTA} & V+E & 85.1 & 89.9 \\
\#06 SDCL~\cite{cao2023event} & V+E & 86.5 & 91.1 \\
\hline 
\#08 TriPro-ReID (Ours) & V+E & 88.4 & 91.1 \\ 
\hline
\toprule [0.5 pt] 
\end{tabular}} 
\end{table}

\subsection{Datasets and Evaluation Metric}  
To evaluate the performance, we conduct a comprehensive benchmark of 10 pedestrian attribute re-identification methods, representing the most important models in the field of pedestrian attribute re-identification. Since there is no available RGB-Event person ReID dataset, Cao et al.~\cite{cao2023event} generate events from MARS~\cite{zheng2016mars}. We adopt this simulated dual-modal MARS RGB-Event dataset (MARS$^*$) and our proposed EvReID dataset for evaluation. Following previous works, we adopt the mean Average Precision (mAP) and Cumulative Matching Characteristics (CMC) at Rank-$\mathcal{K}$ ($\mathcal{K} = 1, 5, 10$) as our evaluation metrics.

\subsection{Implementation Details}  
We use CLIP-ReID~\cite{li2023clip} as our baseline method, and the two modalities adopt independent backbone parameters. Our method consists of three training stages. In the first stage, we only optimize the ID-related context prompts. In the second stage, we introduce 20 randomly initialized cross-modal visual prompts into the RGB branch and generate Event-modality visual prompts via a projection layer. During this stage, only the cross-modal visual prompts and the projection layer are optimized. In the third stage, the text encoder and text context prompts remain frozen. We introduce learnable attribute context prompts and PNAP, and optimize them along with CMP and the visual backbone. SGD is used as the optimizer in the first two stages, while AdamW~\cite{loshchilov2017decoupled} is employed in the final stage. The learning rates are set to $3.5 \times 10^{-3}$, $3.5 \times 10^{-3}$, and $5e-6$, respectively. All training is conducted on an NVIDIA RTX 3090 GPU, with batch sizes of 64, 64, and 8, respectively. More details can be found in our source code.

\begin{table}
\center
\small  
\caption{Ablation study on our proposed EvReID dataset and the public MARS$^*$ dataset. PNAP and CMP denote \textbf{Positive and Negative Attribute Prompts} and \textbf{Cross Modal Prompts}, respectively, where the length of CMP is set as 20.}
\resizebox{0.48\textwidth}{!}{
\label{AblationStudy} 
\raggedright  
\begin{tabular}{c|ccc|cc|cc}
\toprule
\multirow{2}{*}{\textbf{\#Index}} & 
\multirow{2}{*}{\textbf{Base}} & 
\multirow{2}{*}{\textbf{PNAP}} & 
\multirow{2}{*}{\textbf{CMP}} &  
\multicolumn{2}{c|}{\textbf{EvReID}} & 
\multicolumn{2}{c}{\textbf{MARS$^*$}} \\
\cline{5-8}
 & & & & mAP & Rank-1 & mAP & Rank-1 \\
\midrule
1 & \checkmark &  &  &49.2 & 73.0  & 86.8 & 88.7 \\ 
2 & \checkmark & \checkmark &  & 62.3 & 81.1  & 87.2 & 89.9 \\ 
3 & \checkmark &  & \checkmark  & 50.2 & 75.2  & 87.5 & 90.1 \\ 
4 & \checkmark & \checkmark & \checkmark & \textbf{69.3} & \textbf{88.6}  & \textbf{88.4} & \textbf{91.1} \\
\hline\toprule [0.5 pt] 
\end{tabular} }
\end{table}

\begin{table}
\center
\small  
\caption{Comparing different Attribute Prompt settings on EvReID dataset.}
\label{DifferentSettingsOfPNAP}
\resizebox{0.48\textwidth}{!}{
\begin{tabular}{c|cccc}
\hline \toprule [0.5 pt]
\multicolumn{1}{c|}{\textbf{Settings}} & \multicolumn{1}{c}{\textbf{mAP}}  & \multicolumn{1}{c}{\textbf{Rank-1}} & \multicolumn{1}{c}{\textbf{Rank-5}}  & \multicolumn{1}{c}{\textbf{Rank-10}}   \\  
\hline
PNAP  & 69.3 & 88.6 & 94.3 & 95.4 \\
Only Positive  & 54.4 & 78.9 & 89.0 & 91.5 \\ 
w/o Attribute Context Prompts  & 51.4 & 77.4 & 89.6 & 91.2 \\ 
\hline\toprule [0.5 pt] 
\end{tabular} }
\end{table}

\subsection{Comparison with Other SOTA Algorithms} 

\noindent $\bullet$ \textbf{Result on EvReID Dataset.} As shown in Table~\ref{ComparisononEvReIDdataset}, our model achieves impressive mAP, Rank-1, Rank-5, and Rank-10 scores of 69.3/88.6/94.3/95.4, respectively, significantly surpassing existing baseline methods. This substantial performance gain highlights the effectiveness of our dual-branch architecture, which fully exploits the complementary strengths of both RGB and Event modalities. Unlike prior ReID methods, including CNN-based, graph-based, and Transformer-based approaches, our model incorporates a hierarchical cross-modal prompting mechanism that seamlessly integrates multi-modal information. Moreover, attribute prompts further enhance the discriminative capacity of pedestrian features, leading to more robust and distinctive representations. Together, these innovations enable our model to achieve superior accuracy and generalization in person Re-identification tasks.

\noindent $\bullet$ \textbf{Result on MARS$^*$ Dataset.} We further evaluate our TriPro-ReID model on the MARS$^*$ dataset using both RGB and Event modalities. As a large and challenging benchmark for video-based person ReID, MARS$^*$ provides a rigorous testbed for assessing model performance. As shown in Table~\ref{Comparisononpublicdatasets}, our method achieves 88.4\% mAP and 91.1\% Rank-1 accuracy, surpassing all existing state-of-the-art approaches. These results clearly demonstrate the effectiveness of our framework in capturing discriminative spatiotemporal representations, significantly enhancing RGB-Event based person Re-ID performance.

\subsection{Ablation Studies}

\noindent $\bullet$ \textbf{Effect of Key Components.} To evaluate the effectiveness of each proposed component, we conduct ablation studies on the EvReID and MARS$^*$ datasets, as shown in Table~\ref{AblationStudy}. Starting from the base model (Index 1), we observe that introducing the Positive-Negative Attribute Prompt (PNAP) significantly improves performance on both datasets (Index 2). Notably, on EvReID, PNAP increases the mAP from 49.2\% to 62.3\% and Rank-1 from 73.0\% to 81.1\%. We attribute this remarkable gain to the degraded RGB modality in EvReID, which limits the effectiveness of traditional visual features. In such cases, injecting semantic attribute information greatly enriches the feature representation and compensates for the loss of visual cues. On MARS$^*$, although the improvement brought by PNAP is more modest, it still demonstrates the general applicability of attribute prompts in well-illuminated settings.
Adding the Cross Modal Prompts (CMP) alone (Index 3) also contributes positively compared to the base model, yielding consistent improvements on both datasets. CMP enhances the mAP by 1.0\% on EvReID and 0.7\% on MARS$^*$, validating its role in facilitating RGB-Event interaction and capturing modality-complementary information.
When both PNAP and CMP are integrated (Index 4), the full model achieves the best performance. On EvReID, mAP and Rank-1 reach 69.3\% and 88.6\%, respectively, while on MARS$^*$, the model attains 88.4\% mAP and 91.1\% Rank-1. These results demonstrate that PNAP and CMP are complementary modules, and their combination effectively boosts representation learning in both modality-degraded and modality-rich scenarios.

\noindent $\bullet$ \textbf{Effect of Different Attribute Prompt Settings.} Table~\ref{DifferentSettingsOfPNAP} presents a comparative analysis of different attribute prompt configurations on the EvReID dataset. The complete PNAP setting, which includes both positive and negative attribute prompts with contextual information, achieves the best performance, reaching 69.3\% mAP and 88.6\% Rank-1 accuracy. This confirms the effectiveness of bidirectional semantic supervision in enhancing identity discrimination. When using only positive attribute prompts, the performance drops to 54.4\% mAP and 78.9\% Rank-1. This is mainly because some attributes may be shared by multiple individuals, potentially causing identity confusion. The addition of negative prompts further refines semantic boundaries and helps suppress irrelevant or misleading features, resulting in more robust representation learning. Furthermore, removing attribute context prompts entirely leads to a further drop in performance (51.1\% mAP and 77.4\% Rank-1). This suggests that the contextual tokens play a crucial role in enhancing the expressiveness of attribute semantics. Without them, the prompts become isolated keywords lacking sufficient semantic richness, which weakens the alignment between textual descriptions and visual representations. Therefore, context prompts are essential for providing richer and more discriminative semantic guidance.

\begin{table}

\center
\small  
\caption{Comparing different Depths and Lengths of Cross-modal Prompt on EvReID dataset.} 
\label{DepthAndLengthOfCMP}
\setlength{\tabcolsep}{1mm}{
\begin{tabular}{cc|cccc}
\hline \toprule [0.5 pt]
\multicolumn{2}{c|}{\textbf{Length}}  & \multicolumn{1}{c}{\textbf{mAP}}  & \multicolumn{1}{c}{\textbf{Rank-1}} & \multicolumn{1}{c}{\textbf{Rank-5}}  & \multicolumn{1}{c}{\textbf{Rank-10}}    \\  
\hline
& 40 & 62.4 & 81.4 & 90.9 & 93.9  \\
& 20 & 69.3 & 88.6 & 94.3 & 95.4 \\
& 10 & 69.8 & 87.1 & 92.8 & 95.4  \\ 
\hline \toprule [0.5 pt] 
\multicolumn{2}{c|}{\textbf{Depth}}  & \multicolumn{1}{c}{\textbf{mAP}}  & \multicolumn{1}{c}{\textbf{Rank-1}} & \multicolumn{1}{c}{\textbf{Rank-5}}  & \multicolumn{1}{c}{\textbf{Rank-10}}    \\
\hline 
    & 12 & 69.3 & 88.6 & 94.3 & 95.4 \\
    & 6  & 66.7 & 85.8 & 93.4 & 95.3\\
    & 3  & 66.2 & 84.9 & 93.7 & 95.3\\
\hline\toprule [0.5 pt] 
\end{tabular} }
\end{table}

\begin{figure*}
\centering
\includegraphics[width=0.9\textwidth]{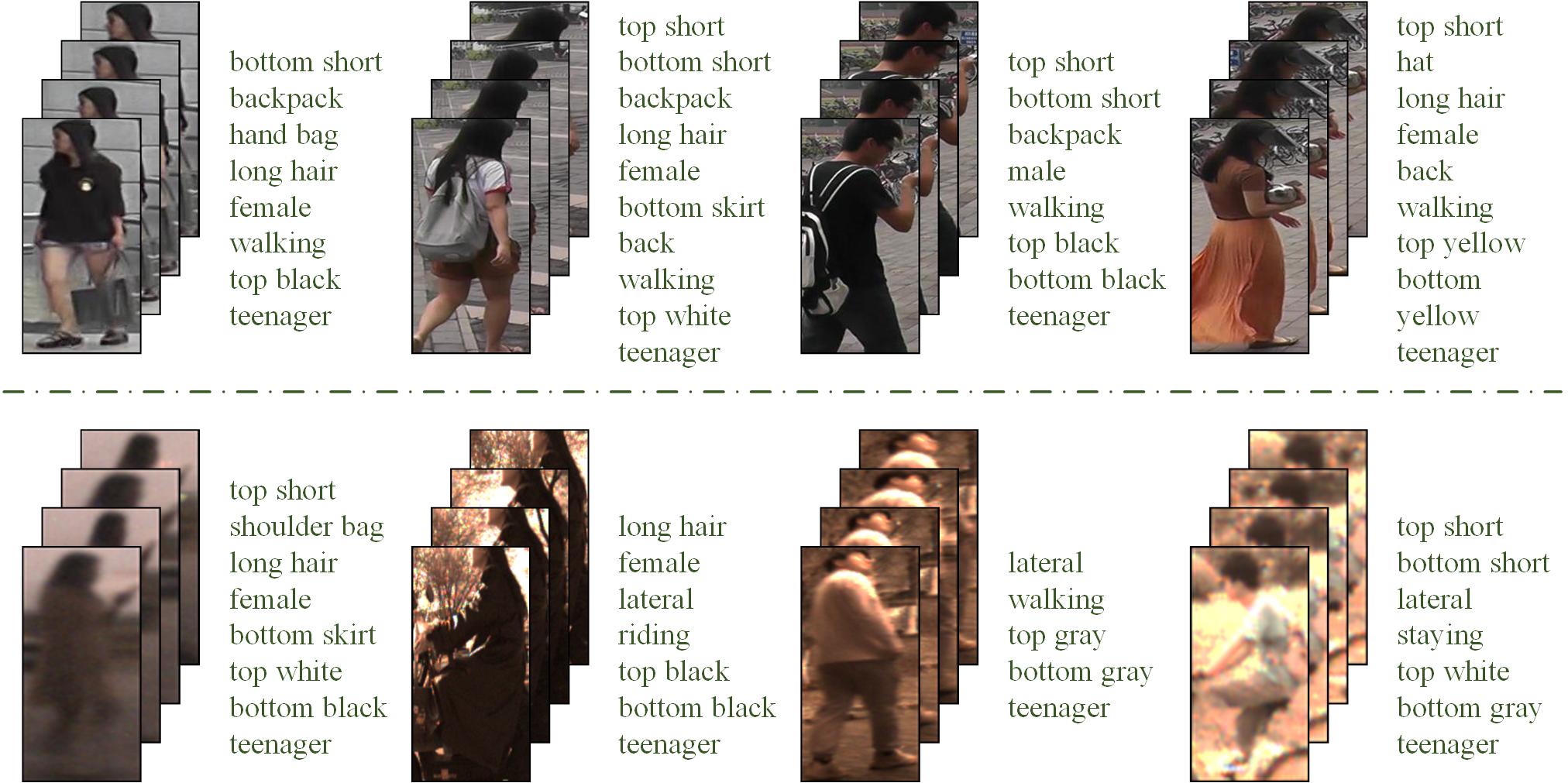} 
\caption{Visualization of pedestrian attribute prediction results. The first row shows the predictions from the MARS dataset, while the second row displays the predictions from the EvReID dataset.}
\label{AttrPresentation}
\end{figure*}

\noindent $\bullet$ \textbf{Effect of Different Depth and Length of CMP.} We investigate the influence of the Cross Modal Prompt (CMP)'s length and depth on the EvReID dataset. As shown in Table~\ref{DepthAndLengthOfCMP}, varying these configurations leads to distinct changes in performance.
For prompt length, we observe that shorter lengths yield better results. Reducing the CMP length from 40 to 20 leads to improvement in mAP (from 62.4\% to 69.3\%) and Rank-1 accuracy (from 81.4\% to 88.6\%). Further shortening the length to 10 achieves the best performance, with 69.8\% mAP and 87.1\% Rank-1. This trend suggests that overly long prompts may introduce redundant or noisy information, while compact prompts encourage more focused and effective cross-modal interactions. For prompt depth, while a depth of 12 yields strong results (69.3\% mAP, 88.6\% Rank-1), reducing it to 6 and 3 causes nearly 3 score drops in accuracy.

\begin{table}
\center
\small  
\caption{Comparing different Projectors of Cross-Modal Prompt on EvReID dataset, where the length of CMP is set as 20.}
\label{DifferentProjectorsForCMP}
\setlength{\tabcolsep}{1mm}{
\begin{tabular}{c|cccc}
\hline \toprule [0.5 pt]
\multicolumn{1}{c|}{\textbf{Settings}} & \multicolumn{1}{c}{\textbf{mAP}}  & \multicolumn{1}{c}{\textbf{Rank-1}} & \multicolumn{1}{c}{\textbf{Rank-5}}  & \multicolumn{1}{c}{\textbf{Rank-10}}  \\  
\hline
    FC & 69.3 & 88.6 & 94.3 & 95.4 \\
    Adapters & 69.7 & 89.7 & 94.3 & 94.3 \\
    None & 68.8 & 86.3 & 94.7 & 95.1  \\
\hline \toprule [0.5 pt] 
\end{tabular} }
\end{table}

\begin{figure}[!htp]
\centering
\includegraphics[width=0.48\textwidth]{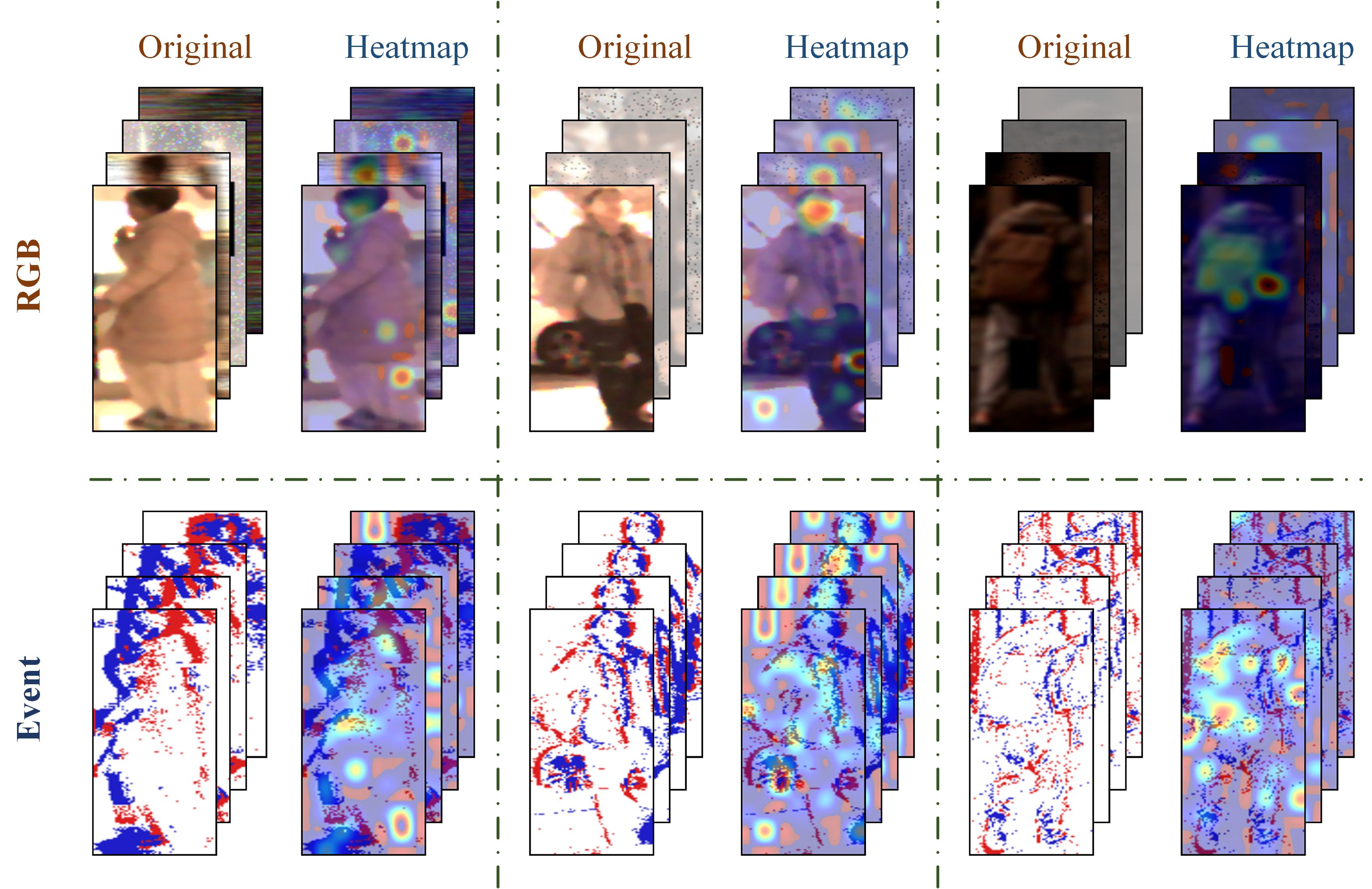} 
\caption{Visualization of pedestrian heatmap results. The first row shows the heatmap for the RGB modality, while the second row displays the corresponding heatmap for the Event modality.}
\label{HeatmapPresentation}
\end{figure}

\noindent $\bullet$ \textbf{Effect of Different Projectors for CMP.} We study the impact of different projection strategies in the design of Cross-Modal Prompts by implementing three variants, including using a fully connected (FC) layer, an adapter module, and a direct mapping approach without projection. As reported in Table~\ref{DifferentProjectorsForCMP}, the adapter-based strategy achieves the best performance, with a mAP of 69.7\% and Rank-1 accuracy of 89.7\%, outperforming both the FC-based (69.3\% / 88.6\%) and the non-projection setting (68.8\% / 86.3\%). 

\begin{table}
\center
\small  
\caption{Comparing different  Cross Modal Type settings on MARS$^*$ dataset.} 
\label{DirectionForCMP}
\setlength{\tabcolsep}{1mm}{
\begin{tabular}{c|cc}
\hline \toprule [0.5 pt]
\multicolumn{1}{c|}{\textbf{Settings}} & \multicolumn{1}{c}{\textbf{mAP}}  & \multicolumn{1}{c}{\textbf{Rank-1}}  \\  
\hline
    RGB $\to$ Event     & 88.4 & 91.1  \\
    Event $\to$ RGB     & 87.9 & 91.1 \\
    Bi-direction        & 87.4 & 90.3 \\
\hline\toprule [0.5 pt] 
\end{tabular} }
\end{table}

\noindent $\bullet$ \textbf{Effect of Different CMP Settings.} To evaluate the influence of different interaction directions in the Cross-Modal Prompt (CMP) module, we compare three representative configurations: unidirectional interaction from RGB to event modality, unidirectional interaction from event to RGB, and bidirectional interaction. Experimental results on the MARS$^*$ dataset, presented in Table~\ref{DirectionForCMP}, indicate that guiding the event branch using prompts generated from the RGB modality achieves the best performance, with 88.4\% mAP and 91.1\% Rank-1 accuracy.

\subsection{Visualization}

\noindent $\bullet$ \textbf{Results of Attributes Prediction.} As shown in Fig.~\ref{AttrPresentation}, we visualize the attribute prediction results on MARS and EvReID datasets. Although the predictions are not entirely comprehensive or perfectly accurate, the pre-trained attribute recognition model is able to capture most of the salient pedestrian attributes. These include clothing types, accessories, and coarse physical characteristics. Such coverage provides sufficiently rich semantic cues, which serve as effective prompts for identity representation learning in our framework.

\noindent $\bullet$ \textbf{Discriminative Attention Maps.} In Fig.~\ref{HeatmapPresentation}, we visualize the learned feature maps to highlight how our network effectively focuses on discriminative, fine-grained details of pedestrians. The attention maps emphasize critical body parts, which are essential for accurate person re-identification, especially under occlusions or pose variations. Our dual-modality approach, combining RGB and event stream data, further enhances this fine-grained focus. RGB provides rich appearance features, while the event stream captures motion and temporal changes. By jointly attending to both modalities, the network integrates complementary information, improving feature representation and alignment between appearance and motion, which boosts performance.

\begin{figure}
\centering
\includegraphics[width=0.48\textwidth]{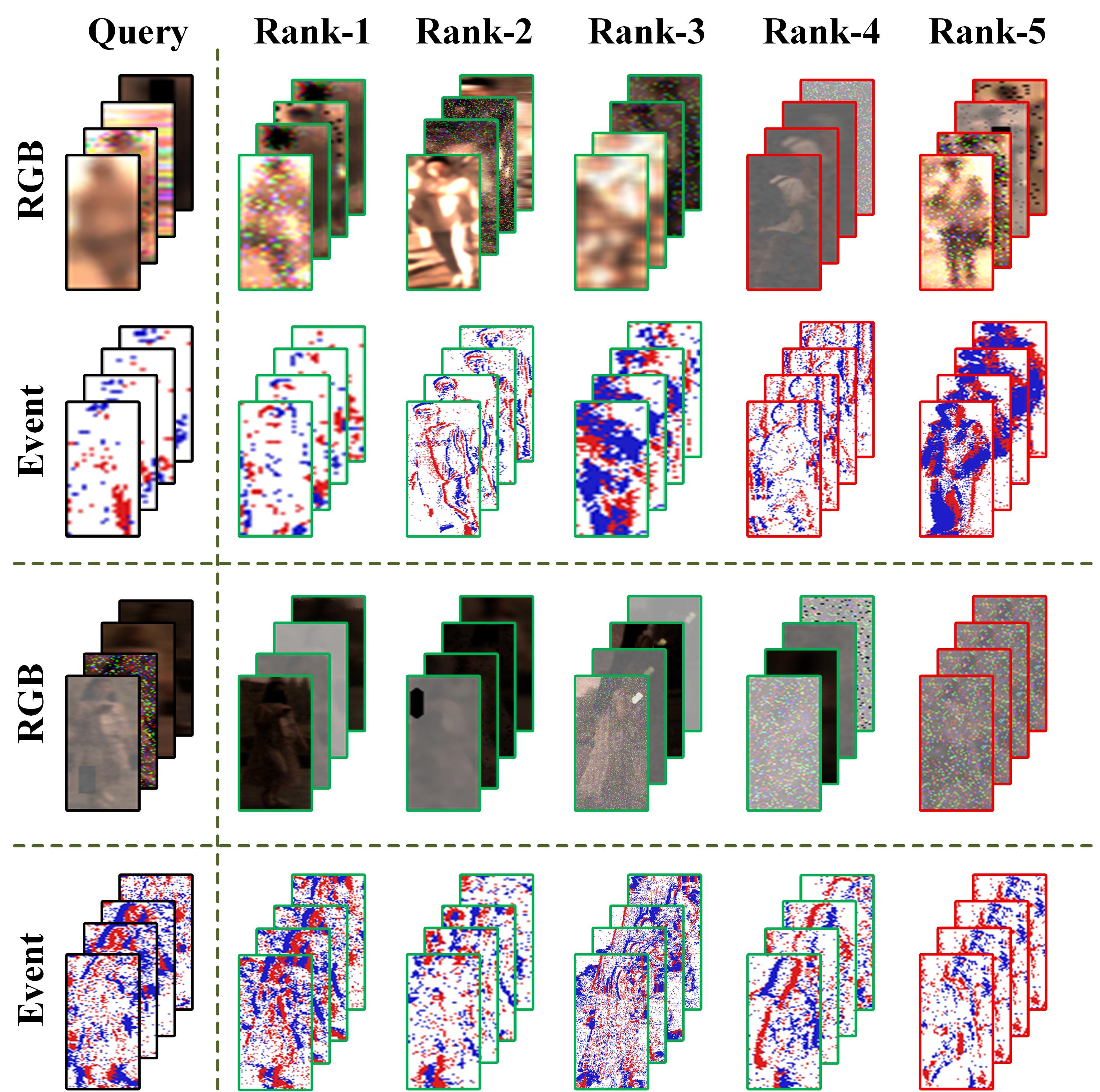} 
\caption{Visualization of the rank list of person re-identification.} 
\label{RankList}
\end{figure}

\noindent $\bullet$ \textbf{Visualization of Rank List.} Fig.~\ref{RankList} shows the rank lists from TriPro-ReID, demonstrating its ability to produce accurate rankings and validate its effectiveness. We present the top-5 ranked results for each query, highlighting the model's precision in ranking and its effectiveness in identifying the correct matches.

\subsection{Limitation Analysis} 
Although the proposed TriPro-ReID achieves promising results on RGB-Event person re-identification, it still has some limitations. First, the Positive-Negative Attribute Prompt (PNAP) relies on a pre-trained attribute recognition model, whose accuracy and generalization may be limited. Incorrect predictions can introduce noise into the prompts and affect feature learning. Moreover, attribute labels are often coarse and shared among different identities, which reduces the discriminative ability of the prompts. Second, the three-stage training pipeline increases overall complexity. Each stage involves specific objectives and model freezing or unfreezing, requiring careful tuning and longer training time. Finally, the use of large vision-language models like CLIP leads to high memory and computation costs, which may hinder deployment on resource-limited platforms.

\section{Conclusion and Future Works}   \label{sec::conclusion}
This paper presents TriPro-ReID, a novel approach for RGB-Event-based person re-identification, which integrates an attribute-guided framework. Our method demonstrates superior performance, addressing key challenges in existing dual-modality person ReID. Furthermore, we introduce a new dataset, EvReID, designed to foster progress in this domain by providing a comprehensive benchmark that spans diverse seasons, scenes, and lighting conditions. In addition, we retrain 15 state-of-the-art methods on this dataset, contributing to the ongoing development of RGB-Event-based person re-identification.
In the future, we plan to expand the scale of the dataset and investigate the impact of temporal dynamics on model performance. Additionally, the use of more advanced large foundation models~\cite{wang2023MMPTMSurvey} will be explored to enhance the overall accuracy.

{
    \small
    \bibliographystyle{ieeenat_fullname}
    \bibliography{main}
}


\end{document}